\documentclass[final]{cvpr}

\usepackage{times}
\usepackage{epsfig}
\usepackage{graphicx}
\usepackage{amsmath}
\usepackage{amssymb}
\usepackage{algpseudocode}

\usepackage[ruled,linesnumbered]{algorithm2e}
\usepackage{multirow}
\usepackage{array}
\usepackage{booktabs}
\usepackage{graphicx}
\usepackage{verbatimbox}
\usepackage{bm}

\usepackage[pagebackref=true,breaklinks=true,colorlinks,bookmarks=false]{hyperref}

\def\cvprPaperID{5182} 
\def\confYear{CVPR 2021}

\begin{document}

\title{Generalizable Person Re-identification with Relevance-aware Mixture of Experts}

\author{{Yongxing Dai{$^{1}$}} \qquad Xiaotong Li{$^{1}$} \qquad Jun Liu{$^{2}$} \qquad Zekun Tong{$^{3}$} \qquad Ling-Yu Duan{$^{1,4}$}\thanks{Corresponding Author.} \\
	\normalsize
	$^{1}$\ National Engineering Lab for Video Technology, Peking University, Beijing, China \\
	\normalsize $^{2}$\ Singapore University of Technology and Design, Singapore \\
	\normalsize $^{3}$\ National University of Singapore, Singapore~~$^{4}$Peng Cheng Laboratory, Shenzhen, China \\
	\normalsize
	{\tt\small \{yongxingdai, lingyu\}@pku.edu.cn, lixiaotong@stu.pku.edu.cn} \\
	{\tt\small jun\_liu@sutd.edu.sg, zekuntong@u.nus.edu}
	}

\maketitle

\thispagestyle{empty}
\begin{abstract}
Domain generalizable (DG) person re-identification (ReID) is a challenging problem because we cannot access any unseen target domain data during training. Almost all the existing DG ReID methods follow the same pipeline where they use a hybrid dataset from multiple source domains for training, and then directly apply the trained model to the unseen target domains for testing. These methods often neglect individual source domains' discriminative characteristics and their relevances w.r.t. the unseen target domains, though both of which can be leveraged to help the model's generalization. To handle the above two issues, we propose a novel method called the relevance-aware mixture of experts (RaMoE), using an effective voting-based mixture mechanism to dynamically leverage source domains' diverse characteristics to improve the model's generalization. Specifically, we propose a decorrelation loss to make the source domain networks (experts) keep the diversity and discriminability of individual domains' characteristics. Besides, we design a voting network to adaptively integrate all the experts' features into the more generalizable aggregated features with domain relevance. Considering the target domains' invisibility during training, we propose a novel learning-to-learn algorithm combined with our relation alignment loss to update the voting network. Extensive experiments demonstrate that our proposed RaMoE outperforms the state-of-the-art methods. 
\end{abstract}

\begin{figure}[t]
\centering
\includegraphics[width=0.9\linewidth]{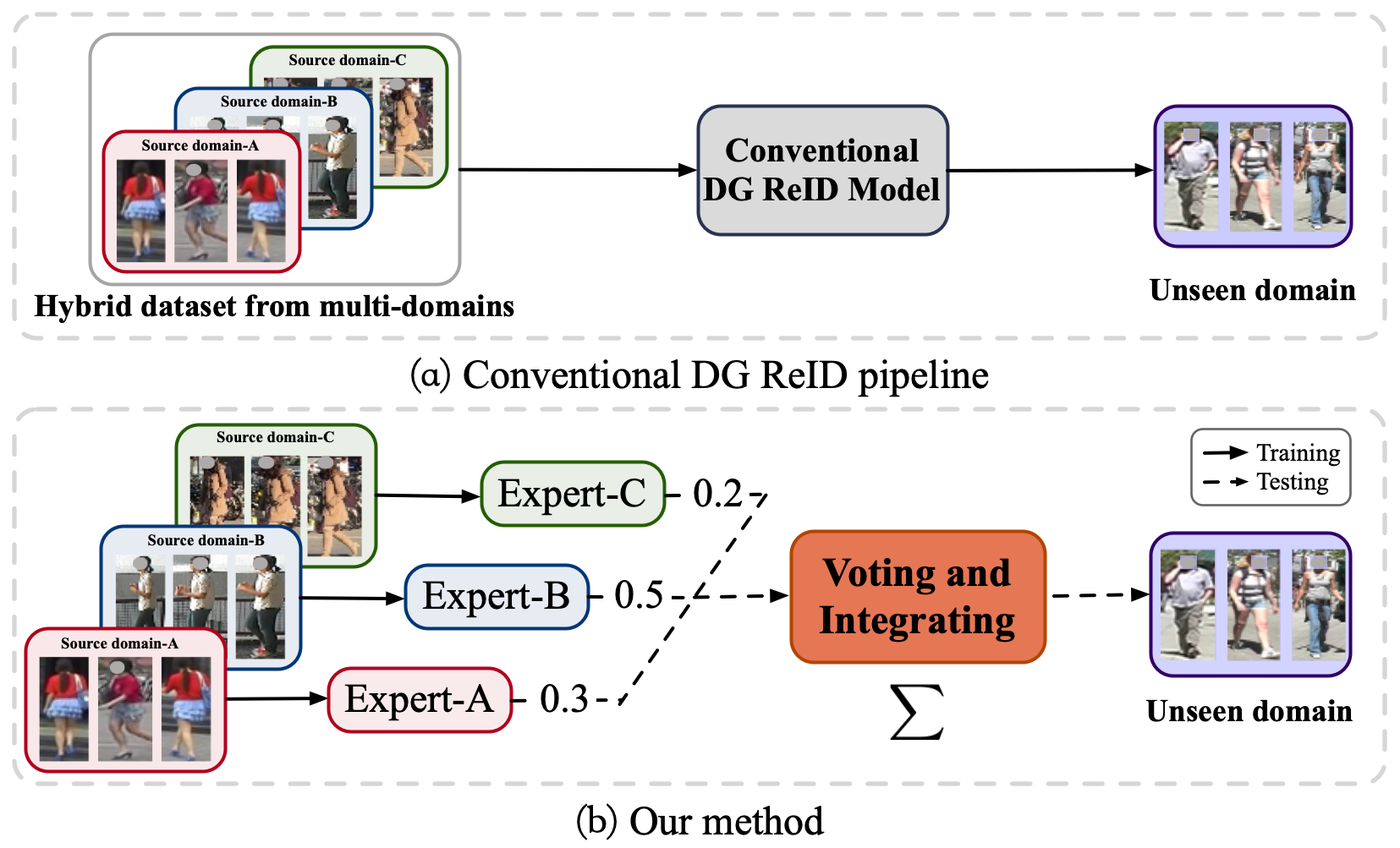}
\caption{
Differences between our method and the conventional DG ReID pipeline. (a) Conventional DG ReID methods generally train a single model on the hybrid dataset from multi-source domains and then apply the trained model to the unseen target domain for testing, which neglects individual domains' discriminative characteristics and target domain's relevance w.r.t. source domains.
(b) Our method leverages the complementary information provided by all the source domain networks (also termed as ``domain experts''). In testing, we integrate features obtained by source domain experts into an adaptive voting process based on the unseen target domain's relevance w.r.t. source domains.
}
   
\label{fig:intro}
\vspace{-0.6cm}

\end{figure}

\section{Introduction}

In recent years, the research on person re-identification (ReID) has been appealing to academia and industry. The goal of ReID is to identify a person across different camera views. 
Many works on fully supervised ReID \cite{sun2019learning,hou2019interaction,chen2019abd} have achieved quite promising performances when training and testing under the same domain (dataset). 
However, when applying these well-trained ReID models to other domains, the performance often drops significantly because of the domain biases \cite{wei2018person}. 
To tackle this problem, some researchers have studied unsupervised domain adaptation (UDA) methods \cite{zhong2019invariance,fu2019self,zhai2020ad,ge2020mutual,chendeep,dai2020dual}, which utilize the unlabeled target data to finetune and adapt the source-trained model to the target domain. 
However, existing UDA ReID methods are often not powerful enough to deal with practical application scenarios, because it is sometimes hard to collect target domain training data and time-consuming to finetune the model on these unlabeled samples. 
As a result, domain generalizable (DG) ReID \cite{song2019generalizable,jia2019frustratingly,jin2020style} has been appealing to researchers recently. Generally, DG ReID methods utilize labeled data from multiple source domains to learn a generalizable model for new unseen target domains, without using any target domain data for training. To obtain more generalizable models for unseen target domains, we are devoted to the problem of DG ReID in this paper.

Almost all the existing DG ReID methods \cite{song2019generalizable,jia2019frustratingly,jin2020style} follow the same pipeline, where they collect all source domain data into a hybrid dataset and train a single model on it, as shown in Fig.~\ref{fig:intro}~\textcolor{red}{(a)}. During testing, they usually use the same well-trained model to extract features for any unseen target domain. However, there can be two potential problems in such a pipeline: 
(1) They learn a common feature space for different domains, which may neglect individual domains' discriminative characteristics. Such diverse domain-specific characteristics have been shown to be able to provide complementary information for better generalization on target domains, as mentioned in \cite{d2018domain,guo2018multi,zhou2020domain}.
(2) Conventional DG ReID methods often ignore the specific target domain's inherent relevance w.r.t. different source domains.
They are difficult to generalize the model to the unseen target domain because the model trained on the more relevant source domains can provide more discriminative and meaningful information than those less relevant domains. 
However, such relevance is often not explicitly considered by existing works ~\cite{song2019generalizable,jia2019frustratingly,jin2020style}.

Recently, works \cite{shazeer2017outrageously,guo2018multi} on the mixture of experts~\cite{jacobs1991adaptive} (MoE) show that MoE can improve the overall model's capability by mixing multiple networks (\textit{i.e.,} leveraging expterts' complementary information) with a voting procedure. 
Inspired by this, we propose a novel approach called Relevance-aware Mixture of Experts (RaMoE), as shown in Fig.~\ref{fig:intro}~\textcolor{red}{(b)}, to handle the above two issues (\textit{i.e.,} complementary information and domain relevance).
We argue that, instead of learning a single model on the hybrid domains, we can train a domain-specific network (domain expert) for each source domain to exploit individual domains' discriminative and powerful characteristics.
Thus, these domain experts' mixture can keep source domains' diversity and provide rich complementary information, improving the generalization on target domains.
Subsequently, we propose an adaptive voting network to calculate the unseen target domain's relevance w.r.t. all source domains. Based on the domain relevance, we can adaptively integrate those source experts' features into the aggregated features by voting. The voting network will assign the more relevant domain experts with higher weights. Thus, those more relevant experts will provide more complementary information to improve the aggregated features' generalizability on the target domain.  

Specifically, in our RaMoE method, we propose a decorrelation loss to encourage source domain experts to keep their domains' diverse characteristics, and thus they can provide complementary and discriminative information.
Such a decorrelation loss is implemented by minimizing the correlation among the source domain experts because the lower correlation among experts will bring about more complementary information, as mentioned in \cite{breiman2001random,opitz2018deep}.  
Because the target domain is totally unseen during training in DG ReID, it is challenging for the adaptive voting network to well learn the target domain's correct relevance w.r.t. source domains. Inspired by meta-learning (learning-to-learn) that can improve the model's generalization \cite{li2018learning,dou2019domain,guo2020learning} for the unseen target domains in an episodic training paradigm, we propose a novel learning-to-learn algorithm to learn our adaptive voting network. At the beginning of each episodic training iteration, we randomly split source domains into the meta-train (simulated ``source domains'') and the meta-test (simulated ``unseen target domains'') to simulate the adaptive voting procedure for the unseen target domain. During each episodic training iteration, the meta-test first obtains the relevance w.r.t. the meta-train using the adaptive voting network. The meta-test can then get two kinds of features: one is the features extracted by the meta-test domain expert, and the other is the aggregated features integrated from multiple meta-train domain experts with the relevance.
We propose the relation alignment loss to push the aggregated features to be as discriminative as the features extracted by the meta-test expert.
As a result, our RaMoE method can generate very discriminative and generalizable aggregated features for the unseen target domains by adaptively integrating diverse domain experts with the domain relevance. 

Our major contributions can be summarized as follows:
(1) We propose a novel RaMoE method to tackle the problem of DG ReID by exploiting source domains' complementary information and their relevance w.r.t. the unseen target domain.
(2) We propose the decorrelation loss to keep source domains' diversity and encourage source domain experts to provide more complementary and discriminative information.
(3) To make the model more generalizable to target domains, we propose a voting network to adaptively integrate source domain experts' features into the aggregated features. Specially, the adaptive voting network is updated with the relation alignment loss in a novel learning-to-learn way. 
(4) Extensive experiments demonstrate that our method outperforms state-of-the-art DG ReID approaches by a large margin.

To the best of our knowledge, this is the first work that treats DG ReID as a novel mixture-of-experts paradigm via an effective voting-based mixture mechanism.

\begin{figure*}
\begin{center}
\includegraphics[width=\linewidth]{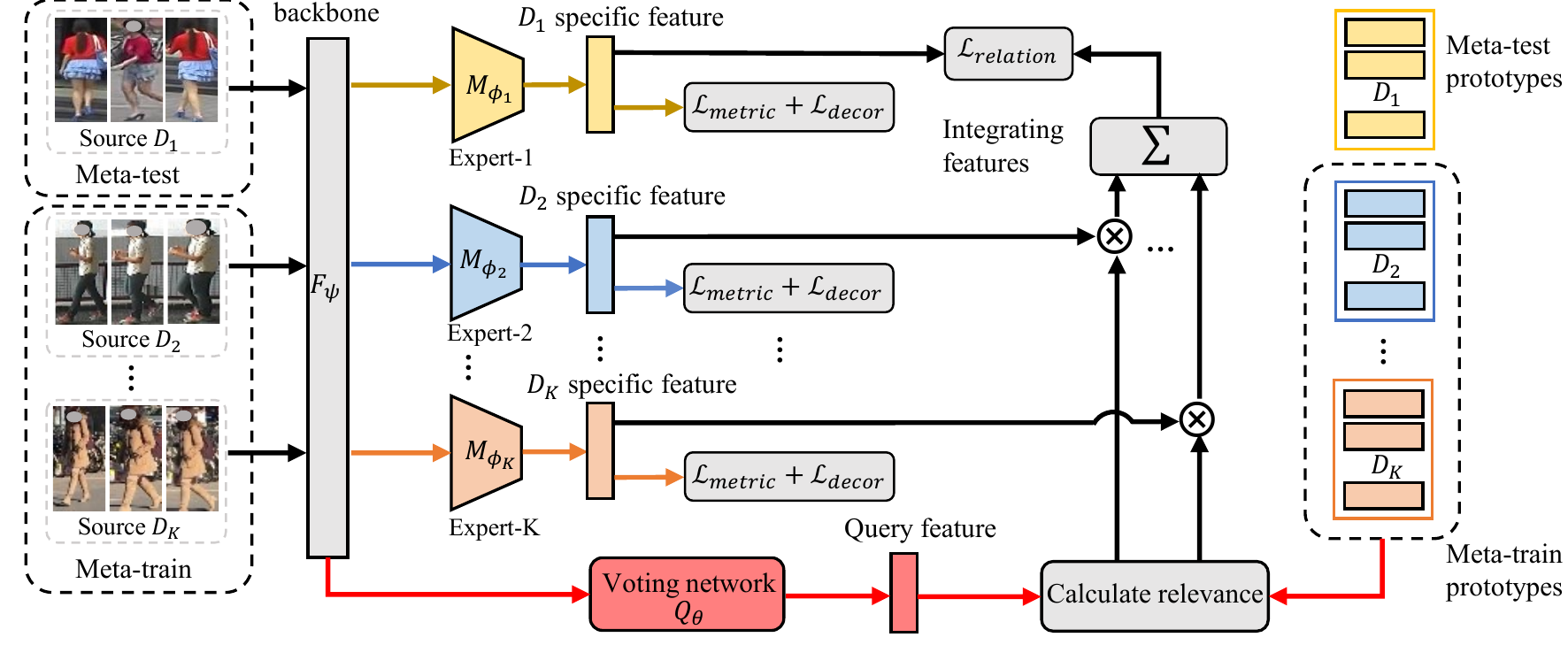}
\end{center}
\vspace{-0.5cm}
\caption{
   Illustration of our method. 
   The $k$-th branch network serves as the expert of the domain $D_{k}$, and it is learned with the metric loss $\mathcal L_{\rm metric}$ and the decorrelation loss $\mathcal L_{\rm decor}$.
   We use the learning-to-learn algorithm combined with the relation alignment loss $\mathcal L_{relation}$ to update the voting network. At each episodic training iteration, we split $K$ domains into the meta-test (\textit{e.g.,} $D_{1}$) and meta-train (\textit{e.g.,} $D_{2}, ..., D_{K}$). A meta-test image can obtain $K$ features (one feature from its own domain expert and $K-1$ features from meta-train domain experts), together with a query feature obtained by the voting network. The meta-test domain's relevance w.r.t. the meta-train can be obtained by calculating the mean similarity between the meta-test query feature and the meta-train prototypes. We can obtain the weighted aggregated feature by adaptively integrating meta-train experts' features with the relevance. ``$\bigotimes$'' is the operation of weighting features with the domain relevance.
   The relation alignment loss is proposed to push the weighted aggregated feature as discriminative as the meta-test domain-specific feature. ``$\sum$'' is the features' operation: concatenation or element-wise summation.
   }
\vspace{-1.6em}
\label{fig:pipeline}
\end{figure*}

\section{Related Work}

\textbf{Person Re-Identification.}
Deep supervised person ReID has made great progress in recent years, including but not limited to deep metric learning \cite{hermans2017defense,chen2017beyond,chen2019deep,sun2020circle}, part-based methods \cite{sun2019learning,li2017learning,suh2018part,guo2019beyond}, and attention network learning \cite{chen2019mixed,chen2019self,zhang2020relation}. 
To handle the problem of domain biases \cite{deng2018image, wei2018person} in ReID, researchers proposed unsupervised domain adaptation (UDA) methods \cite{zhong2019invariance,fu2019self,zhai2020ad,ge2020mutual,chendeep,dai2020dual,ge2020self,zheng2020exploiting,zheng2021group}.
Very recently, researchers started to study the topic of domain generalization (DG) in ReID \cite{song2019generalizable,jia2019frustratingly,jin2020style}, which learns the generalizable ReID models on multi-source domains without using any target training data, and tests on unseen target domains. 
Song \textit{et al.} \cite{song2019generalizable} proposed the problem of domain generalization in ReID and designed the Domain-Invariant Mapping Network combined with a memory bank to learn domain-invariant features. 
Jia \textit{et al.} \cite{jia2019frustratingly} utilized Instance Normalization \cite{ulyanov2016instance} to learn a more generalizable model. 
Jin \textit{et al.} \cite{jin2020style} proposed Style Normalization and Restitution modules to disentangle the identity-relevant and identity-irrelevant features.
Different from all the above DG ReID works, we propose a novel RaMoE method by utilizing individual source domains' diverse and discriminative characteristics and the unseen target domain's relevance w.r.t. source domains in order to adaptively improve the model's generalization on unseen target domains.

\textbf{Domain Generalization.} The goal of general DG is to improve the model generalization in an arbitrary domain for image classification by training from multi-source domains. Existing DG methods can be mainly categorized into three aspects.
(1) Learning domain-invariant features \cite{muandet2013domain,ghifary2015domain,li2018domain}: These methods assume that minimizing the domain discrepancy between multi-source domains can help learn domain-invariant features which are robust for unseen target domains. 
(2) Augmenting source data \cite{shankar2018generalizing,volpi2018generalizing,carlucci2019domain,zhou2020domain,zhou2020learning}: These methods augment the source domain data to increase the domain diversity, thus the source-trained model will be more robust to unseen target domains. 
(3) Optimizing with meta-learning \cite{li2018learning,li2019episodic,dou2019domain}: These methods adopted the episodic training paradigm to split the source domains into meta-train and meta-test to simulate the domain bias, so as to improve the model generalization.
The above general DG methods mainly focus on image classification where the target domains and the source domains share the same label space. Thus, these methods can not be directly applied to the task of DG ReID, since in ReID, the identities/classes of the target domains are usually totally different from source domains.

\textbf{Mixture of Experts.} Jacobs \textit{et al.} \cite{jacobs1991adaptive} first introduced the mixture of experts (MoE). MoE aims to learn a system composed of many separated networks (experts), where each expert learns to handle a subset of the whole dataset. Recently, deep MoE methods have shown their superiority in image recognition \cite{ahmed2016network,gross2017hard,wang2020deep}, machine translation \cite{shazeer2017outrageously}, scene parsing \cite{fu2018moe} and so on. 
Unlike these works, we design a learnable voting network that can be updated with a novel meta-learning algorithm. By integrating all the experts using our designed voting network, we can well leverage the complementary information of those relevant domains' experts to improve the features' generalization in DG ReID.

\section{Methodology}
In this work, we aim to train a group of experts that are capable of learning discriminative features from their individual domains. When facing an unseen target domain, the mixture of these domain experts can be trained to vote based on their relevances w.r.t. the target domain. 
By adaptively integrating all the source experts' features into aggregated features with the relevance, our RaMoE can achieve an optimal generalization performance on the target domain.

\subsection{Overview}
The pipeline of our proposed RaMoE is illustrated in Fig.~\ref{fig:pipeline}. During training, we can access $K$ source domains' labeled datasets ${\bm{D}}=\{ {D}_{k} \}_{k=1}^{K} $, where $D_{k}=\{(x_{n}^{k},y_{n}^{k})\}_{n=1}^{N_{k}}$, $(x_{n}^{k},y_{n}^{k})$ is a labeled sample and $N_{k}$ is the number of labeled images in the $k$-th domain. After the backbone $F_{\psi }$ (\textit{e.g.,} ResNet50), we design $K$ branch networks (termed as ``source domain experts'') $\{M_{\phi_{k}}\}_{k=1}^{K}$, and a voting network $Q_{\theta}$. 
The metric loss $\mathcal L_{\rm metric}$ makes each expert focus on learning its domain-specific features. The decorrelation loss $\mathcal L_{\rm decor}$ is used to keep source domains' diverse characteristics and encourage all the domain experts to provide complementary information. We use the $k$-th domain's class centers $C_{k}=\{c_{l}^{k}\}_{l=1}^{L_{k}}$ as the prototypes to represent the $k$-th domain's characteristics, 
where $L_{k}$ is the number of person identities in the $k$-th domain.

We propose a novel meta-learning algorithm combined with the relation alignment loss $\mathcal L_{\rm relation}$ to update the voting network. $K$ source domains are randomly split into a meta-test domain and $K-1$ meta-train domains at each episodic training iteration. For a meta-test image, it can obtain $K+1$ features, including (1) a feature extracted by the meta-test expert, (2) $K-1$ features extracted by the meta-train experts, (3) a query feature extracted by the voting network. The meta-test domain's relevance w.r.t. the meta-train domains can be calculated by the mean similarity between the query feature and the meta-train domains' prototypes. We can obtain the weighted aggregated feature by integrating $K-1$ meta-train expert features based on their relevance. The relation alignment loss is proposed to push the weighted aggregated feature to be as discriminative as the meta-test feature.

\subsection{Optimizing Domain-specific Experts}
As mentioned in \cite{d2018domain,guo2018multi,zhou2020domain}, exploiting the complementary information of discriminative experts helps improve the overall model's generalization on target domains. Thus, the domain experts should satisfy two properties: discriminability and complementarity. We use the metric loss to improve every domain-specific expert's discriminability. Similar to \cite{opitz2018deep}, we mutually reduce all the domain experts' correlation to improve the complementarity among them. Specifically, we propose a decorrelation loss to decorrelate all these domain experts' features.

\textbf{Metric Loss.} Similar to \cite{luo2019strong}, we use the classification loss $\mathcal L_{\rm cls}$, triplet loss \cite{hermans2017defense} $\mathcal L_{\rm tri}$, and center loss \cite{wen2016discriminative} $\mathcal L_{\rm cent}$ to optimize $K$ domain-specific experts $\{M_{\phi_{k}}\}_{k=1}^{K}$, the domain-specific prototypes $\{C_{k}\}_{k=1}^{K}$, and the backbone network $F_{\psi }$. 
We combine the above metric losses as: 
\vspace{-0.5em}
\begin{equation}
\small
\label{eq:loss_metric}
\mathcal L_{\rm metric}=\mathcal L_{\rm cls}+\mathcal L_{\rm tri}+\lambda \mathcal L_{\rm cent},
\end{equation}
where $\lambda$ (set as $5\times10^{-4}$) is the weighting hyper-parameter. 

\textbf{Decorrelation Loss.}
For an  image $x_{n}^{k}$ (where $n=1,2,...,N_{k}$) from the $k$-th domain, we use all the experts to extract $K$ features $\{m_{n}^{j}\}_{j=1}^{K}$ that are characterized by individual domains, as shown in Fig.~\ref{fig:pipeline}.
To improve the aggregated features' generalization, we encourage these experts to provide more complementary and discriminative information. Specifically, we propose the decorrelation loss by reducing the correlation among different domain experts. We formulate the decorrelation loss as follows:
\vspace{-0.2em}
\begin{equation}
\small
\label{eq:loss_decor}
\mathcal L_{\rm decor}=\frac{1}{N_{k}}\sum_{n=1}^{N_{k}}(\frac{1}{K-1}\sum_{j\neq k}|| m_{n}^{k}\odot m_{n}^{j} ||),
\vspace{-0.2em}
\end{equation}
where features $\{m_{n}^{j}\}_{j=1}^{K}$ are all L2-normalized, $\odot$ means the point-wise product and $\left \| \cdot \right \|$ is the L2-norm of a vector. 

We combine Eq. (\ref{eq:loss_metric}) (\ref{eq:loss_decor}) into the domain loss by: 
\vspace{-0.2em}
\begin{equation}
\small
\label{eq:loss_domain}
\mathcal L_{\rm domain}= \mathcal L_{\rm metric}+ \mathcal L_{\rm decor}.
\vspace{-0.1cm}
\end{equation}
Thus, by alternating $k$ from $1$ to $K$, we can obtain a group of representative and complementary domain experts.

\subsection{Optimizing the Voting Network}
To make the model more generalizable to the unseen target domain, we leverage the specific target domain's relevance w.r.t. all source domains. Specifically, we propose a voting network to calculate the domain relevance adaptively. 
By integrating all the source domain experts' features into a weighted aggregated feature with relevance, we can achieve more generalizable features for an unseen target domain during testing.
Because the target domain data is unavailable during training, we propose a learning-to-learn algorithm to simulate integrating multi-source experts' features with the relevance. The voting network can be updated with a relation alignment loss introduced below. Thus, we can learn a generalizable voting network for an unseen target domain, integrating multi-source experts' features adaptively. Specifically, we split the $K$ source domains into meta-train (simulated ``source domains'') $\bm D_{s}$ including $K-1$ domains, and the meta-test (simulated ``the unseen target domain'') $\bm D_{u}$ including the remaining domain, at every episodic training iteration.

\textbf{Relation Alignment Loss.} As mentioned before,
for a $k$-th domain's image $x_{n}^{k}$ (where $n=1,2,...,N_{k}$), we can obtain $K$ features $\{m_{n}^{j}\}_{j=1}^{K}$ extracted by $K$ experts $\{M_{\phi_{j}}(\cdot ) \}_{j=1}^{K}$, and a query feature $q_{n}^{k}$ extracted by the voting network $Q_{\theta}(\cdot )$, as illustrated in Fig. \ref{fig:pipeline}. 

We use the query feature $q_{n}^{k}$ to calculate the domain relevance score of the $k$-th domain's image $x_{n}^{k}$ w.r.t. the $j$-th domain ($j\neq k$) by:
\vspace{-0.5em}
\begin{equation}
\small
\label{eq:relevance}
    s_{n}^{j}=\frac{1}{L_{j}}\sum_{l=1}^{L_{j}} \big< q_{n}^{k},c_{l}^{j} \big>,
\vspace{-0.5em}
\end{equation}
where $\big <q_{n}^{k},c_{l}^{j} \big> $ is the inner product between the query feature $q_{n}^{k}$ and the $l$-th class prototype $c_{l}^{j}$ (where $l=1,2,...,L_{j}$) in the $j$-th domain. Both $q_{n}^{k}$ and $c_{l}^{j}$ are L2-normalized. 
As a result, we can get the relevance set $\{s_{n}^{j}\}_{j=1,j\neq k}^{K}$ of the image $x_{n}^{k}$ w.r.t. all other $K-1$ domains. Thus, for a $k$-th domain image $x_{n}^{k}$, we can then integrate other $K-1$ irrelevant experts' features $\{m_{n}^{j}\}_{j=1,j\neq k}^{K}$ into the weighted aggregated feature $v_{n}$ with the relevance $s_{n}^{j}$ by:
\vspace{-0.2em}
\begin{equation}
\small
\label{eq:integrating}
    v_{n}=\sum_{j\neq k}\sigma (s_{n}^{j})\cdot m_{n}^{j},
\end{equation}
where $\sigma(\cdot)$ is the non-linear function (\textit{e.g.,} sigmoid or softmax) to normalize the relevance between 0 and 1.

Softmax-triplet function \cite{ge2020mutual, ye2020distilling} has been shown to be a powerful tool to measure the metric relationship in the feature space (\textit{i.e.,} inter-sample discriminability). Thus we use it to measure the metric relationship of the weighted aggregated feature $v_{n}$ as below:
\begin{equation}
\small
\label{eq:softmax-triplet}
    R(v_{n})=\frac{\mathrm{exp}(\left \| v_{n}-v_{n}^{+}\right \|)}{\mathrm{exp}(\left \| v_{n}-v_{n}^{+}\right \|)+\mathrm{exp}(\left \| v_{n}-v_{n}^{-}\right \|)},
\end{equation}
where $R(\cdot)\in[0,1]$, $\left \| \cdot \right \|$ is the L2-norm of a vector, and $v_{n}^{+}$ and $v_{n}^{-}$ are the selected features of the hardest positive and negative samples within a mini-batch. 
Similarly, for the $k$-th expert's feature $m_{n}^{k}$ we can also use Eq.~(\ref{eq:softmax-triplet}) obtain the metric relationship $R(m_{n}^{k})$. 

Compared with other $K-1$ domain experts, the $k$-th domain expert should be able to generate more discriminative feature for the sample $x_{n}^{k}$, while such metric relationship $R(m_{n}^{k})$ reflects the $k$-th domain-specific discriminative characteristics. Thus, we push the weighted aggregated feature $v_{n}$ to be as discriminative as the $k$-th domain-specific feature $m_{n}^{k}$, and meanwhile, enable the weighted aggregated feature to be characterized by the $k$-th domain, we propose the relation alignment loss below:
\begin{equation}
\small
\label{eq:loss_relation}
    \mathcal L_{\rm relation}=\frac{1}{N_{k}}\sum_{n=1}^{N_{k}}\mathcal L_{bce}(R(v_{n}),R(m_{n}^{k})),
\end{equation}
where $\mathcal L_{\rm bce}$ is the binary cross-entropy loss. By minimizing Eq. (\ref{eq:loss_relation}), the voting network is pushed to learn to produce reliable relevance scores. Thus, the model can learn powerful generalization capabilities for unseen target domains, by exploiting how to integrate source domains.

\begin{algorithm}
\footnotesize
    \caption{Training Procedure of RaMoE}
    \label{alg:algorithm1}
    \KwIn{
    Source domains $ \bm{D}=\{{D}_{k} \}_{k=1}^{K} $; Learning rate hyperparameters $\alpha, \beta, \gamma$; Balance hyperparameter $\eta$; MaxIters; MaxEpochs.
    
    }
    \KwOut{
    Backbone feature extractor $ F_{\psi } $; Domain-specific experts $\{M_{\phi_{k} }\}_{k=1}^{K}$; 
    Prototypes $\{C_{k}\}_{k=1}^{K}$; Voting network $Q_{\theta }$.
    }
    
    // For simplicity, we denote $\mathcal L_{\rm domain}$ and $\mathcal L_{\rm relation}$ as ${\mathcal L_{d}}$ and ${\mathcal L_{r}}$ respectively.
    
    \For{$epoch=1$ to MaxEpochs}
    {
     \For{$iter=1$ to MaxIters}
        {
        Sample $K-1$ domains as meta-train $\bm {D}_{s}$ and the remaining as meta-test $\bm {D}_{u}$;
        
        \textbf{Meta-training:}
        
        Compute losses for $\bm {D}_{s}$: $\mathcal L^{s}=\mathcal L_{d}^{s} + L_{r}^{s} (\theta )$;

        Update the voting network parameters by:
        \qquad \qquad \qquad ${\theta}'\leftarrow \theta - \alpha \nabla_{\theta}  \mathcal L_{r}^{s}(\theta)$;
        
        \textbf{Meta-testing:}
        
        Compute losses for $\bm {D}_{u}$: $\mathcal L^{u}=\mathcal L_{d}^{u}+\mathcal L_{r}^{u} ({\theta}' )$;

        \textbf{Optimizing:}
        
        $\psi \leftarrow \psi - \beta \nabla_{\psi}(\mathcal L_{d}^{s}+\mathcal L_{d}^{u})$;
        
        $(\phi_{s} ,\bm C_{s})\leftarrow (\phi_{s} ,\bm C_{s}) - \beta \nabla_{\phi_{s} ,\bm C_{s}}\mathcal L_{d}^{s}$;
        
        $(\phi_{u} ,\bm C_{u})\leftarrow (\phi_{u} ,\bm C_{u}) - \beta \nabla_{\phi_{u} ,\bm C_{u}}\mathcal L_{d}^{u}$;
        
        \textbf{Meta-optimizing}
        
        $\theta\leftarrow \theta - \gamma ((1-\eta)\nabla_{\theta}  \mathcal L_{r}^{s}(\theta)+ \eta\nabla_{\theta}  \mathcal L_{r}^{u}(\theta'))$;
        }
        }

\end{algorithm}

\textbf{Meta Optimizing.}
Since we cannot access the unseen target domain samples, we design a meta-learning scheme to optimize the above losses.
At the \textbf{meta-training} stage, we use the meta-train $\bm D_{s}$ to compute the domain loss with Eq.~(\ref{eq:loss_domain}) and the relation alignment loss with Eq.~(\ref{eq:loss_relation}) as:
\begin{equation}
\label{eq:loss_meta_train}
\small
    \mathcal L^{s}=\mathcal L_{\rm domain}^{s}( \bm D_{s};\psi ,\phi_{s} ,\bm C_{s})+ \mathcal L_{\rm relation}^{s} (\bm D_{s};\psi ,\phi_{s}, \bm C_{s}, \theta ),
\end{equation}
where $\psi$ is the parameter of the backbone, $\phi_{s}$ is the parameter of the domain-specific experts of $\bm D_{s}$, $\bm C_{s}$ is the prototypes set of $\bm D_{s}$, and $\theta$ is the parameter of the voting network. 
Similar to \cite{wen2016discriminative}, prototypes can be updated with the center loss in Eq. (\ref{eq:loss_metric}).
Next, the updated parameters of the voting network is obtained by: ${\theta}'\leftarrow \theta - \alpha \nabla_{\theta}  \mathcal L_{\rm relation}^{s}(\theta)$, where $\alpha$ is the learning rate hyper-parameter.
At the \textbf{meta-testing} stage, we use the meta-test $\bm D_{u}$ to compute the domain loss and relation alignment loss with Eq.~(\ref{eq:loss_domain})~(\ref{eq:loss_relation}), which is formulated as follows:
\begin{equation}
\small
    \mathcal L^{u}=\mathcal L_{\rm domain}^{u}( \bm D_{u};\psi ,\phi_{u}, \bm C_{u})+\mathcal L_{\rm relation}^{u} (\bm D_{u};\psi ,\phi_{u}, \bm C_{u}, {\theta}' ),
\end{equation}
where $\phi_{u}$ is the parameter of the $\bm D_{u}$ expert, $\bm C_{u}$ is the prototypes set of $\bm D_{u}$, and ${\theta}'$ is the updated parameter with Eq.~(\ref{eq:loss_meta_train}). 
At the \textbf{meta-optimizing} stage, we optimize the voting network with the second-order gradient as follows:
\begin{equation}
\small
    \theta\leftarrow \theta - \gamma ((1-\eta)\nabla_{\theta}  \mathcal L_{\rm relation}^{s}(\theta)+ \eta\nabla_{\theta}  \mathcal L_{\rm relation}^{u}(\theta')),
\end{equation}
where $\gamma$ is the learning rate and $\eta$ (set as 0.5) is the hyper-parameter to balance the gradient of meta-train and meta-test. The overall training procedure is shown in Alg.~\ref{alg:algorithm1}.

\subsection{Testing Procedure}
\label{sec:testing}
During testing, for the unseen target domain dataset consisting of $N$ samples $\{x_{n}\}_{n=1}^{N}$, we use Eq. (\ref{eq:relevance}) to obtain the relevance of each target sample $x_{n}$ w.r.t. all $K$ source domains, \textit{i.e.,} $\{s_{n}^{k}\}_{k=1}^{K}$. Then, we can obtain the relevance of the unseen target domain w.r.t. the $k$-th source domain by $s^{k}=\frac{1}{N}\sum_{n=1}^{N}s_{n}^{k}$. 
Each target sample $x_{n}$ can 
achieve $K$ features $\{m_{n}^{k}\}_{k=1}^{K}$ using $K$ domain experts. Similar to Eq.~(\ref{eq:integrating}), we adaptively integrate all $K$ source domains' features with the relevance $\{s_{n}^{k}\}_{k=1}^{K}$ by:

\begin{equation}
\small
    \label{eq:test-integrating}
    v_{n}=\sum_{k=1}^{K}\sigma (s^{k})\cdot m_{n}^{k} \ \ ,
\end{equation}
where the weighted aggregated features $\{v_{n}\}_{n=1}^{N}$ are all L2-normalized for evaluating.

\section{Experiments}

\subsection{Implementation Details}
We use ResNet50 \cite{he2016deep} pretrained on ImageNet as our backbone. Similar to \cite{luo2019strong}, the last residual layer's stride size is set as 1. After the global pooling layer we add an Embedding layer (\textit{i.e.,} FC: 2048d$\to$512d) followed by batch normalization (BN) to get the ReID feature. The identity classifier (Classifier) followed by softmax function is added after BN to optimize with the classification loss. The above network is the structure of our \textbf{\textit{Baseline}}. 
For efficiency, in our method, we make all the source domains share the same backbone and add a branch network (expert) for each source domain. Specifically, the structure of every domain expert is Embedding$\to$BN$\to$Classifier.
The voting network can be easily implemented with FC$\to$ReLU$\to$BN, where FC is 2048d$\to$512d.
We resize the person image size to 256 × 128. For data augmentation, we perform random cropping, random flipping, and color jittering. Similar to~\cite{jin2020style}, we discard random erasing (REA) because REA will degenerate the cross-domain ReID performance \cite{luo2019strong}. 
The batch size is set to 64, including 16 identities and four images per identity. 
For our \textbf{\textit{Baseline}}, we combine all the source domains into a hybrid dataset and only use the metric loss $\mathcal L_{\rm metric}$ for training. In our \textbf{RaMoE} method, we sample each source domain evenly at every training iteration. 
We optimize the model with the Adam optimizer.
We train the model for 120 epochs and use the warmup strategy in the first ten epochs. The learning rate (\textit{i.e.,} $\alpha, \beta, \gamma$ in Alg.~\ref{alg:algorithm1}) is initialized as $3.5 \times 10^{-4}$ and divided by 10 at the 40th and 70th epochs respectively. We conduct all the experiments with PyTorch and train the model on four 1080Ti GPUs. The training and testing are efficient in our multi-head RaMoE method where the training and inference time of each batch are 0.708s and 0.312s respectively (batch size is 64).

\subsection{Datasets and Evaluation Settings.}
\textbf{Datasets and Evaluation Metrics.} Following the previous works \cite{song2019generalizable,jia2019frustratingly,jin2020style} on DG ReID, we conduct our experiments on the public ReID or PearsonSearch datasets, including Market1501 \cite{zheng2015scalable}, DukeMTMC-reID \cite{zheng2017unlabeled}, CUHK02 \cite{li2013locally}, CUHK03 \cite{li2014deepreid}, MSMT17 \cite{wei2018person}, CUHK-SYSU \cite{xiao2016end}, and four small ReID datasets including PRID \cite{hirzer2011person}, GRID \cite{loy2010time}, VIPeR \cite{gray2008viewpoint}, and iLIDs \cite{wei2009associating}. 
For CUHK03, we use the ``labelled'' dataset for training and adopt the protocol used in \cite{zhong2017re} for testing.
For simplicity, in the next sections we denote Market1501 as M, DukeMTMC-reID as D, CUHK02 as C2, CUHK03 as C3, MSMT17 as MT, and CUHK-SYSU as CS. We use the mean average precision (mAP) and Cumulative Matching Characteristics (CMC) for evaluation.

\begin{table}[tp]
\footnotesize
\caption{Different evaluation protocols. The leave-one-out setting for M+D+C3+MT means selecting one domain for testing and the remaining three domains for training.}
\vspace{-0.2cm}
\label{tab:protocols}
\begin{center}
\begin{tabular}{|c|l|l|}
\hline
Setting   & \multicolumn{1}{c|}{Training Data} & \multicolumn{1}{c|}{Testing Data}                                                   \\ \hline
Protocol-1 & M+D+C2+C3+CS                       & \multirow{2}{*}{\begin{tabular}[c]{@{}l@{}}PRID, GRID,\\ VIPeR, iLIDs\end{tabular}} \\ \cline{1-2}
Protocol-2 & M+D+C3+MT                          &                                                                                     \\ \hline
Protocol-3 & \multicolumn{2}{c|}{Leave-one-out for M+D+C3+MT}                                                                         \\ \hline
\end{tabular}
\end{center}
\vspace{-3.5em}
\end{table}

\textbf{Evaluation Protocols.} There exist two evaluation protocols for DG ReID, as shown in Tab. \ref{tab:protocols}. Under the setting of Protoco1-1 \cite{song2019generalizable}, all the images in these datasets M+D+C2+C3+CS (including the training and testing sets) are used for training. Four small ReID datasets (\textit{i.e.,} PRID, GRID, VIPeR, and iLIDs) are tested respectively, where the final performances of these small ReID datasets are evaluated on the average of 10 repeated random splits of gallery and probe sets.
Under Protocol-2~\cite{jin2020style}, all the images in M+D+C3+MT (including the training and testing sets) are used for training and the testing sets are the same as Protocol-1. However, two disadvantages may lie in Protocol-1 and Protocol-2: 
(1) Compared with the existing ReID datasets, the number of images per identity in the CS dataset is much smaller, which will limit the learning of discriminative ReID features.
(2) The images' quality of the four small ReID datasets is low.
The small datasets' performances can not correctly evaluate the model's generalizability in real scenarios, where the latter needs to be evaluated on large-scale datasets.
As a result, we set a new protocol (\textit{i.e.,} Protocol-3 in Tab.~\ref{tab:protocols}) of the leave-one-out setting for the existing large-scale public datasets M+D+C3+MT. Specifically, the leave-one-out setting of M+D+C3+MT is selecting one domain from M+D+C3+MT for testing (only the testing set in this domain) and all the remaining domains for training (including the training and testing sets).
\begin{table*}[htp]
\footnotesize
\caption{Comparison with state-of-the-arts methods in DG ReID under the setting of protocol-1 and protocol-2. We report the performances of the methods marked by `` * '' from \cite{song2019generalizable}. The best results are highlighted with bold.} 
\vspace{-0.3cm}
\label{tab:sota}
\begin{center}

\begin{tabular}{l|l|c|cccccccc}
\toprule[1pt]
\multicolumn{1}{c|}{\multirow{2}{*}{Setting}} & \multicolumn{1}{c|}{\multirow{2}{*}{Method}} & \multicolumn{1}{c|}{\multirow{2}{*}{Reference}} & \multicolumn{2}{c}{Target: PRID} & \multicolumn{2}{c}{Target: GRID} & \multicolumn{2}{c}{Target: VIPeR} & \multicolumn{2}{c}{Target: iLIDs} \\
\multicolumn{1}{c|}{} & \multicolumn{1}{c|}{} & \multicolumn{1}{c|}{} & mAP & Rank-1 & mAP & Rank-1 & mAP & Rank-1 & mAP & Rank-1 \\ \hline
\multirow{11}{*}{Protocol-1} & Agg\_Align* \cite{zhang2017alignedreid} &arXiv 2017  & 25.5 & 17.2 & 24.7 & 15.9 & 52.9 & 42.8 & 74.7 & 63.8 \\
 & Reptile* \cite{nichol2018first} &arXiv 2018  & 26.9 & 17.9 & 23.0 & 16.2 & 31.3 & 22.1 & 67.1 & 56.0 \\
 & CrossGrad* \cite{shankar2018generalizing} &ICLR 2018  & 28.2 & 18.8 & 16.0 & 8.96 & 30.4 & 20.9 & 61.3 & 49.7 \\
 & Agg\_PCB* \cite{sun2019learning} &TPAMI 2019  & 32.0 & 21.5 & 44.7 & 36.0 & 45.4 & 38.1 & 73.9 & 66.7 \\
 & MLDG* \cite{li2018learning} &AAAI 2018  & 35.4 & 24.0 & 23.6 & 15.8 & 33.5 & 23.5 & 65.2 & 53.8 \\
 & PPA* \cite{qiao2018few} &CVPR 2018  & 45.3 & 31.9 & 38.0 & 26.9 & 54.5 & 45.1 & 72.7 & 64.5 \\
 & DIMN* \cite{song2019generalizable} &CVPR 2019  & 52.0 & 39.2 & 41.1 & 29.3 & 60.1 & 51.2 & 78.4 & 70.2 \\
 & SNR \cite{jin2020style} &CVPR 2020  & 66.5 & 52.1 & 47.7 & 40.2 & 61.3 & 52.9 & 89.9 & 84.1 \\ \cline{2-11} 
 & \textbf{\textit{Baseline}} & \multicolumn{1}{c|}{\multirow{2}{*}{CVPR 2021}}  & 60.4 & 47.3 & 49.0 & 39.4 & 58.0 & 49.2 & 84.0 & 77.3 \\
 & \textbf{RaMoE (Ours)} & \multicolumn{1}{c|}{}  & \textbf{67.3} & \textbf{57.7} & \textbf{54.2} & \textbf{46.8} & \textbf{64.6} & \textbf{56.6} & \textbf{90.2} & \textbf{85.0} \\ \hline
\multirow{3}{*}{Protocol-2} & SNR \cite{jin2020style} &CVPR 2020  & 60.0 & 49.0 & 41.3 & 30.4 & 65.0 & 55.1 & 91.9 & 87.0 \\ \cline{2-11} 
 & \textbf{\textit{Baseline}} &\multicolumn{1}{c|}{\multirow{2}{*}{CVPR 2021}}  & 58.9 & 47.2 & 47.7 & 38.1 & 63.8 & 54.7 & 89.2 & 84.2 \\
 & \textbf{RaMoE (Ours)} &  & \textbf{66.8} & \textbf{56.9} & \textbf{53.9} & \textbf{43.4} & \textbf{72.2} & \textbf{63.4} & \textbf{92.3} & \textbf{88.4} \\ 
\bottomrule[1pt]
\end{tabular}
\end{center}
\vspace{-2.8em}
\end{table*}

\begin{table}[htp]
\footnotesize
\caption{Comparisons under the setting of protocol-3.}
\label{tab:protocol3}
\centering
\begin{tabular}{l|cccc}
\toprule[1pt]
\multicolumn{1}{c|}{Target: Market} & mAP & Rank-1 & Rank-5 & Rank-10 \\ \hline
\textbf{\textit{Baseline}} &49.9  &75.4  &86.9  &91.0  \\
\textbf{RaMoE (Ours)} 
&\textbf{56.5}  
&\textbf{82.0}  
&\textbf{91.4}
&\textbf{94.4}  \\ \hline
\multicolumn{1}{c|}{Target: Duke} & mAP & Rank-1 & Rank-5 & Rank-10 \\ \hline
\textbf{\textit{Baseline}}  &49.4  &65.8  &79.0  &83.9  \\
\textbf{RaMoE (Ours)} 
&\textbf{56.9}  
&\textbf{73.6}  
&\textbf{85.3}  
&\textbf{88.4}  \\ \hline
\multicolumn{1}{c|}{Target: CUHK03} & mAP & Rank-1 & Rank-5 & Rank-10 \\ \hline
\textbf{\textit{Baseline}} & 32.6 & 32.9 &52.9  &63.6  \\
\textbf{RaMoE (Ours)} 
& \textbf{35.5} 
& \textbf{36.6} 
&\textbf{54.3}  
&\textbf{64.6}  \\ \hline
\multicolumn{1}{c|}{Target: MSMT17} & mAP & Rank-1 & Rank-5 & Rank-10 \\ \hline
\textbf{\textit{Baseline}} & 9.9 & 24.5 &35.4  &40.9  \\
\textbf{RaMoE (Ours)} 
& \textbf{13.5} 
& \textbf{34.1} 
& \textbf{46.0}  
& \textbf{51.8}  \\ 
\bottomrule[1pt]
\end{tabular}
\vspace{-0.5cm}
\end{table}

\subsection{Comparison with the State-of-the-Arts}
Our proposed RaMoE can outperform the state-of-the-arts methods by a large margin in the task of Domain Generalization (DG) ReID, as shown in Tab. \ref{tab:sota}. 
The \textbf{\textit{Baseline}} method is training on the hybrid dataset including all source domains with only the metric loss $\mathcal L_{\rm metric}$.

\textbf{Comparison with DG ReID methods under the Protocol-1 and Protocol-2.} We compare our method with the existing DG ReID methods under two different evaluation protocols. 
All the other methods directly apply the model trained on source domains to the unseen target domain without considering the domain relevance. Compared with them, our RaMoE can outperform them significantly.

\textbf{Comparison under the Protocol-3.} We compare our proposed \textbf{RaMoE} with the \textbf{\textit{Baseline}} method under the protocol-3 in Tab. \ref{tab:protocol3}.  
The performances on these large-scale ReID datasets have shown our method's superiority in integrating source domains' characteristics adaptively for better domain generalization.
\begin{table*}[htp]
\centering
\caption{We study ablation studies on individual components of our method under the Protocol-2. Voting means learning the voting network with meta-learning by $\mathcal L_{\rm relation}$ and decorrelation means decorrelating source domain experts by $\mathcal L_{\rm decor}$. Expert-M/D/C3/MT means using the feature extracted by Market/Duke/CUHK03/MSMT17 domain expert. Experts-ensemble means concatenating source domain experts' features directly without learning the domain relevance.}
\label{tab:ablation}
\footnotesize
\begin{tabular}{l|cccccccc}
\toprule[1pt]

\multicolumn{1}{c|}{\multirow{2}{*}{Method}} & \multicolumn{2}{c}{Target: PRID} & \multicolumn{2}{c}{Target: GRID} & \multicolumn{2}{c}{Target: VIPeR} & \multicolumn{2}{c}{Target: iLIDs} \\
\multicolumn{1}{c|}{}                        & mAP           & Rank-1           & mAP           & Rank-1           & mAP            & Rank-1           & mAP            & Rank-1           \\ \hline
w/o voting (Expert-M) &62.2 &53.4  &50.4 &39.8 &66.1 &56.9 &87.9 &82.7 \\
w/o voting (Expert-D) &61.6 &51.6 &48.4 &38.5 &67.0 &57.5 &90.3 &85.5 \\
w/o voting (Expert-C3) &62.5 &53.7 &51.0 &41.4 &68.5 &59.7 &88.7 &84.0 \\
w/o voting (Expert-MT) &63.6 &54.9 &49.9 &40.0 &66.9 &57.3 &89.0 &84.5 \\ \hline
w/o voting (Experts-ensemble) &65.1 &56.1 &52.3 &42.2 &70.6 &61.6 &91.4 &86.7 \\ 
w/o decorrelation  &66.0 &55.6 &53.2 &42.9 &71.3 &62.8 &91.2 &87.0 \\ 
\textbf{RaMoE (Ours)} &\textbf{66.8} &\textbf{56.9} &\textbf{53.9} &\textbf{43.4} &\textbf{72.2} &\textbf{63.4} &\textbf{92.3} &\textbf{88.4}  \\ 
\bottomrule[1pt]
\end{tabular}
\vspace{-1.5em}
\end{table*}

\subsection{Ablation Study}

\begin{table}[tp]
\footnotesize
\caption{Evaluating on different non-linear functions $\sigma(\cdot)$ and feature integrating types under the setting of Protocol-2.}
\label{tab:integrating-types}
\centering
\addvbuffer[0cm 0cm]{
\begin{tabular}{cc|cc|cc}
\toprule[1pt]
 \multicolumn{2}{c|}{Non-linear $\sigma(\cdot)$} & \multicolumn{2}{c|}{Integrating type} &
 \multicolumn{2}{c}{Target: GRID} \\
softmax & sigmoid & concat & sum &mAP & R1 \\ \hline
\checkmark  &  &\checkmark  &  &53.9 &43.4   \\
  &\checkmark  &\checkmark  &  &53.7 &43.3   \\
\checkmark  &  &  &\checkmark &52.3 &41.2  \\
  &\checkmark  &  &\checkmark &51.9 &40.9  \\ 
\bottomrule[1pt]
\end{tabular}}
\vspace{-1em}
\end{table}

\begin{table}[tp]
\footnotesize
\caption{Evaluation of mAP within source domains. 
}
\label{tab:test-on-source}
\centering
\begin{tabular}{l|cccc}
\toprule[1pt]
Method & M & D & C3 & MT \\ \hline
Single-source Baseline & 81.8 & 71.6 & 62.0 & 46.6 \\
Multi-source Baseline & 82.6 & 74.4 & 64.3 & 48.0 \\
RaMoE (Ours) & 83.8 & 74.6 & 65.6 & 49.1 \\
\bottomrule[1pt]
\end{tabular}
\vspace{-1.8em}
\end{table}

\textbf{Effectiveness of the domain decorrelation.} 
We propose the decorrelation loss $\mathcal L_{\rm decor}$ to encourage source domain experts to keep their diverse and discriminative characteristics. Thus, integrating these experts can provide complementary information to improve the aggregated features' generalization. 
As shown in Tab.~\ref{tab:ablation}, our method outperforms ours w/o decorrelation by 1.3\% in Rank-1 on PRID. If learning source experts without the decorrelation loss, 
the experts will provide less complementary information and thus reduce the generalization of the aggregated features.

\textbf{Effectiveness of the voting network.} The voting network learned with meta-learning can adaptively provide the relevance of the target domain w.r.t. source domains, making those more relevant source domains provide more complementary information to improve the generalization of the weighted aggregated features. As shown in Tab.~\ref{tab:ablation}, our method outperforms ours w/o voting (Experts-ensemble) by 1.7\%, 1.6\%, 1.6\%, 0.9\% in mAP on PRID, GRID, VIPeR, and iLIDs respectively. Experts-ensemble means that the relevance of the target domain w.r.t. source domains is set 1, and all the experts' features are directly concatenated into the ensemble features. However, our method uses the domain relevance to integrate adaptively. Take the performances on iLIDs as an example, the Expert-M performs worst compared with other three experts (\textit{i.e.,} Expert-D/C3/MT) and the Expert-D performs best. Though directly mixing all these experts (\textit{i.e.,} Experts-ensemble) can bring about great performance gain, the methods w/o voting is inferior to our RaMoE significantly. It can demonstrate that the voting mechanism using the domain relevance can adaptively leverage those more relevant experts' complementary information and alleviate the influence of those less relevant experts.

\textbf{Can individual domain experts provide complementary information to improve the features' generalization?}
We can keep all the source domains' diverse and discriminative characteristics using the decorrelation loss. Thus, all the source domain experts are encouraged to provide more complementary information.
As shown in Tab.~\ref{tab:ablation}, 
almost all the experts (\textit{i.e.,} Expert-M/D/C3/MT) do not perform very well on different target domains. However, when integrating these experts' features, the aggregated features are superior to those extracted by individual experts. Thus, we can improve the overall features' generalization for unseen target domains by leveraging individual source domains' complementary information. 

\textbf{How to integrate different source domain features?} As shown in Tab. \ref{tab:integrating-types}, 
we compare different combinations of non-linear functions $\sigma(\cdot)$ and feature integrating types. The results show that the types of the non-linear function $\sigma(\cdot)$ in Eq. (\ref{eq:integrating}) will not bring about significant performance fluctuations. When concatenating features obtained by different source domain experts, the performance is better than summing features along with the corresponding dimensions, because the type of concatenating will keep more information about the different feature dimensions. Thus, we choose the combination of ``softmax'' and ``concat'' to integrate source domains' features for our method in all the experiments.

\subsection{Extension}

\textbf{Evaluation on source domains.}
We use M, D, C3, and MT as source datasets, where only their training sets are used to train, and their testing sets are only used to test. 
In Tab.~\ref{tab:test-on-source}, Single-source Baseline means training and testing on the single domain; Multi-source Baseline means training on a hybrid dataset of all domains and testing
on each domain separately. Comparing with them, the accuracy of RaMoE on source datasets does not drop but increases.

\textbf{Extension to the online setting.} We can easily extend our testing procedure to a more practical setting where the query set samples are given online, \textit{i.e.,} only the gallery samples are used to calculate the domain relevance in testing. We compare our method with this online setting in Tab.~\ref{tab:gallery4relevance} and there is only very negligible performance drop when only using gallery to calculate the domain relevance.

\begin{figure}[tp]
\centering
\includegraphics[width=0.8\linewidth]{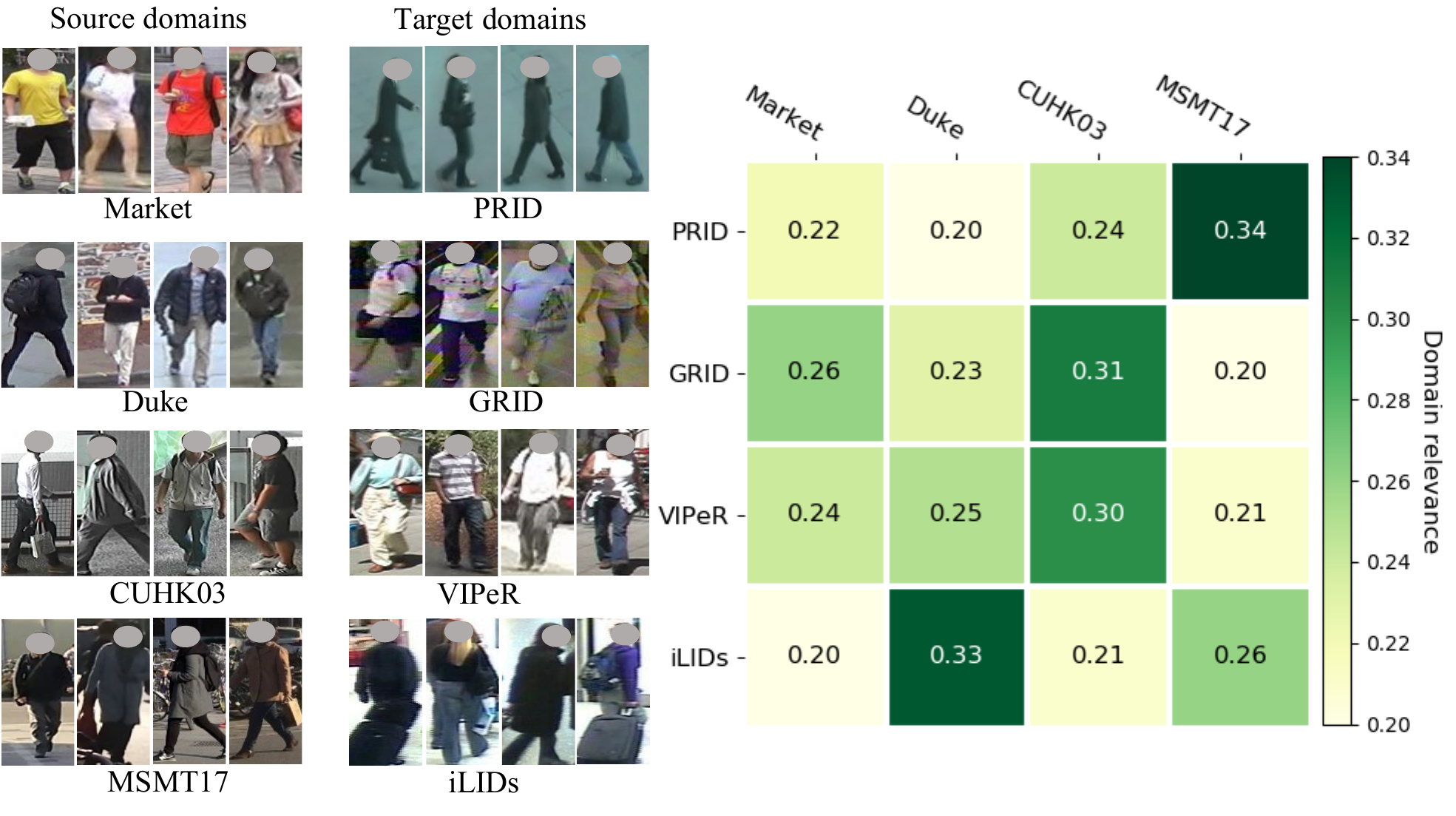}
\caption{Visualization on the domain relevance.}
\label{fig:relevance}
\vspace{-0.6cm}
\end{figure}

\textbf{Visualization.} As shown in Fig.~\ref{fig:relevance}, we visualize the domain relevance between the target domains (\textit{i.e.,} PRID, GRID, VIPeR, and iLIDs) w.r.t. the source domains
(\textit{i.e.,} Market, Duke, CUHK03, and MSMT17), where the domain relevance is calculated with the manner mentioned in Sec.~\ref{sec:testing}. In Fig.~\ref{fig:relevance}, we can see that the unseen target domain's relevance w.r.t. all the source domains are different, and there exist some more relevant source domains for the unseen target domain. For the unseen target domain dataset iLIDs, its style is more similar to MSMT17 and Duke, and thus their relevances are higher than the other two datasets.

\begin{table}[tp]
\footnotesize
\caption{Evaluation of mAP on different kinds of calculating the domain relevances $\{s^{k}\}$ in testing under the setting of Protocol-3.}
\label{tab:gallery4relevance}
\centering
\begin{tabular}{l|cccc}
\toprule[1pt]
Samples used for calculating $\{s^{k}\}$ & M & D & C3 & MT \\ \hline
All samples (Ours) & 56.5 & 56.9 & 35.5 & 13.5 \\
Only samples in gallery & 56.3 & 56.8 & 35.4 & 13.5 \\ 
\bottomrule[1pt]
\end{tabular}
\vspace{-1em}
\end{table}

\section{Conclusion}
This paper proposes a novel approach called Relevance-aware Mixture of Experts (RaMoE) to tackle the problem of domain generalizable person ReID (DG ReID). By considering both the source domains' individual discriminative characteristics and the relevance of the unseen target domain w.r.t. source domains, we can obtain more generalizable features adaptively for the unseen target domain in DG ReID. Specifically, we propose the decorrelation loss to keep source domains' diverse and discriminative characteristics. Thus, these experts can provide  more complementary information to improve the aggregated features' generalization. To obtain more accurate domain relevance of the unseen target domain w.r.t. source domains, we propose the voting network learned with the relation alignment loss in a meta-learning way. Extensive experiments show the effectiveness of our proposed RaMoE method.

\noindent \textbf{Acknowledgements:} 
This work was supported by the National Natural Science Foundation of China under Grant 62088102, and in part by the PKU-NTU Joint Research Institute (JRI) sponsored by a donation from the Ng Teng Fong Charitable Foundation, and was partially supported by SUTD Project PIE-SGP-Al2020-02. 

{\small
\bibliographystyle{ieee_fullname}
\bibliography{egbib}
}

\end{document}